# Novelty-focused R&D landscaping using transformer and local outlier factor

*Jaewoong Choi*

Computational Science Research Center, Korea Institute of Science and Technology, 5 Hwarang-ro 14-gil, Seongbuk-gu, Seoul 02792, Republic of Korea

[*]Correspondence to Jaewoong Choi (jwoongchoi407@gmail.com)

**Abstract**

While numerous studies have explored the field of research and development (R&D) landscaping, the preponderance of these investigations has emphasized predictive analysis based on R&D outcomes, specifically patents, and academic literature. However, the value of research proposals and novelty analysis has seldom been addressed. This study proposes a systematic approach to constructing and navigating the R&D landscape that can be utilized to guide organizations to respond in a reproducible and timely manner to the challenges presented by increasing number of research proposals. At the heart of the proposed approach is the composite use of the transformer-based language model and the local outlier factor (LOF). The semantic meaning of the research proposals is captured with our further-trained transformers, thereby constructing a comprehensive R&D landscape. Subsequently, the novelty of the newly selected research proposals within the annual landscape is quantified on a numerical scale utilizing the LOF by assessing the dissimilarity of each proposal to others preceding and within the same year. A case study examining research proposals in the energy and resource sector in South Korea is presented. The systematic process and quantitative outcomes are expected to be useful decision-support tools, providing future insights regarding R&D planning and roadmapping.

# 1. Introduction

Monitoring R&D activities within national and organizational innovation systems has become increasingly critical, as it supports the development of strategic, forward-looking R&D planning and roadmapping without redundancy [1, 2]. By analyzing past and present R&D activities and forecasting near-future trends, policymakers and strategists can shape the direction of innovation more effectively. However, as modern R&D ecosystems grow more complex and dynamic, traditional expert-based approaches such as Delphi and surveys have become not only time-consuming and labor-intensive but also less effective in providing timely, actionable insights [3, 4]. As a result, policymakers and practitioners have called for reliable, data-driven methods to effectively monitor and navigate the evolving R&D landscape. In response, researchers have proposed data-driven approaches using scientific publications and technical documents to systematically explore and predict the R&D landscape.

The implications of data-driven R&D landscape analysis can vary significantly depending on the data source used. Specifically, three primary data sources, patents, academic publications, and web data, offer distinct perspectives. First, patents are widely regarded by researchers as reliable sources of technical information for R&D landscape analysis [5-7]. Patents capture the tangible outcomes of R&D efforts and often represent technologies, functionalities, or systems that have been developed or are nearing market readiness. Second, academic publications focus on the research stage, presenting methodologies, theoretical models, and experimental findings, typically before industrial applications [8, 9]. Because both patents and academic papers are subject to review and approval processes, a time lag exists between the initial occurrence of R&D activities and the point at which they can be analyzed. Third, web data offer forward-looking insights, with implications that vary across platforms. For instance, Wikipedia provides collaboratively generated and verified content on various technologies associated through hyperlinks, enabling us to obtain early insights into technological convergence [10]. Similarly, technology foresight websites reflect expert opinions on emerging trends and anticipated shifts [11], making them valuable for tracking early indicators of future changes, such as weak signals, despite inherent uncertainties.

To effectively map the current R&D landscape and project its near future direction, advancements in data sources and methodologies are essential. Conventional data sources such as patents and research publications predominantly represent R&D outcomes, whereas web-based data lacks sufficient relevance to R&D activities. We introduced research proposals that capture researchers' immediate R&D intentions as a new source for R&D landscaping. Notably, newly selected research proposals, which have undergone a rigorous review process as new ideas different from ongoing ones, are employed for R&D landscape analysis [12, 13]. Given the potential of innovative research to drive future shifts in the R&D ecosystem, and with novelty recognized as a fundamental catalyst of innovation [14], we center our analysis on assessing novelty within these proposals. However, while R&D proposals serve as valuable data sources, their technical content and terminologies present challenges in systematically evaluating novelty.

As a remedy, we suggest a systematic approach to define the R&D landscape and measure the novelty of R&D proposals through a structured, quantifiable process. The near future of the R&D landscape becomes clearer by identifying novel research proposals composed of new problem-solution pairs, rather than interpreting a set of technical keywords in patents. At the core of the proposed approach are the transformer model and local outlier factor (LOF) technique. The transformer model is employed to interpret domain-specific textual information within research proposals, whereas the LOF quantitatively assesses the novelty of each proposal. Here, novelty is defined as the distinctiveness of newly selected research proposals compared to others before and within the same year, where the semantic meanings of proposal elements such as titles, research objectives, content, and expected outcomes are compared. This novelty metric provides new text-mined insights into R&D monitoring and planning by capturing the novelty of research ideas at the proposal stage.

We applied the proposed approach to 12,243 R&D proposals in South Korea's energy/resource sector between 2010 and 2022. This case study demonstrates the capability of this approach to systematically construct an R&D landscape and identify novel research concepts. We further trained pretrained transformer-based language models on this R&D dataset to achieve a higher level of comprehension tailored to the R&D context. Our findings show that novel proposals statistically outperform others in R&D continuity, scale, and outcomes, providing empirical support for the role of novelty in driving impactful R&D. Although these results do not constitute an absolute validation of our novelty metrics, they suggest that innovation often stems from unique sources of novelty that contribute to successful outcomes. The proposed approach and quantitative outcomes offer a valuable tool for policymakers and strategic planners to enhance decision-making in R&D planning and technology roadmapping.

The remainder of this paper is organized as follows. Section 2 describes the technical background of the transformer model and local outlier factor. Section 3 explains the proposed approach, which is illustrated using the case study in Section 4. Section 5 discusses the academic and practical implications of the proposed approach and Section 6 presents the current limitations and future studies.

## 2. Background
### 2.1. Transformer
Transformer [15], proposed by Google in 2017, is a sequence-to-sequence (seq2seq) architecture that employs self-attention mechanisms, and marked a significant advancement in language modeling. Its main characteristics can be summarized as self-attention mechanism, positional encoding, and multi-head. First, self-attention operates by transforming each input token into three vectors, query (Q), key (K), and value (V), which capture relational information by computing pairwise relevance scores among tokens (Figure 1). This mechanism allows the transformer to establish dependencies between distant tokens without the sequential constraints of prior models, thereby overcoming issues related to information loss in long sequences. Beyond self-attention, the transformer integrates two additional key components: positional encoding and multi-head attention. Positional encoding is applied to convert token positions into vectors using trigonometric functions, which are then added to the token embeddings. The multi-head attention mechanism enhances feature extraction by capturing diverse relationships across multiple low-dimensional subspaces and combining them for richer representation.

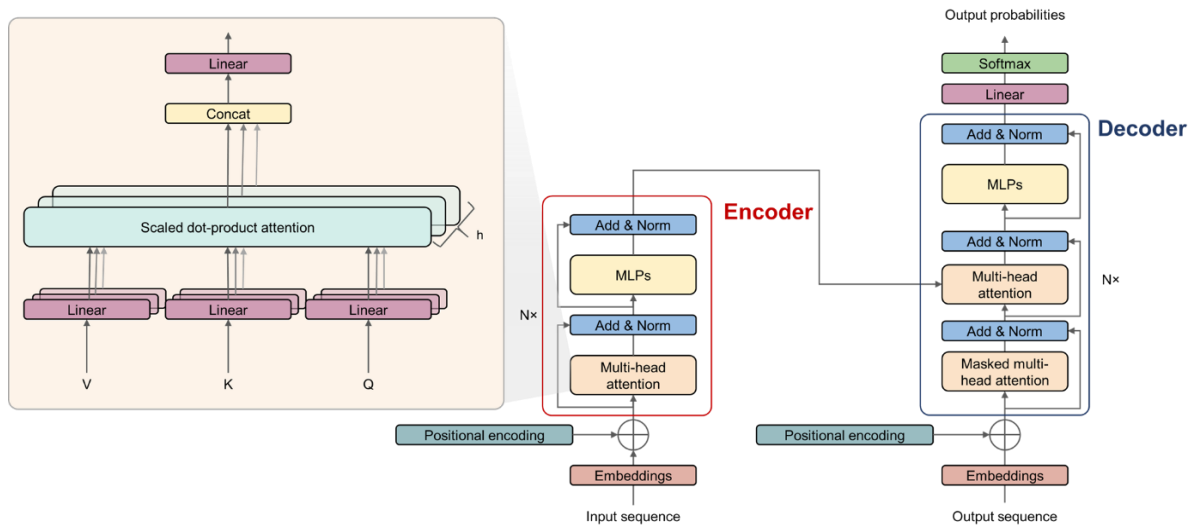

**Figure 1.** Architecture of transformer model

Since the emergence of transformer models, various adaptations have evolved, and are primarily categorized into two types: autoencoding and autoregressive. First, the autoencoding models, notably represented by the bidirectional encoder representations from transformers (BERT) [16], originate from the encoder architecture of the transformer. This encoder simultaneously processes all words in an input sequence, enabling superior contextual understanding and bidirectional information flow. BERT is trained through two primary tasks: masked language modeling, where select words in an input sentence are masked and subsequently predicted, and next sentence prediction, which assesses the relationship between two consecutive sentences. BERT involves extensive pretraining on large corpora, followed by fine-tuning for specific applications. On the other hand, autoregressive models for text generation are developed, based on the decoder of the transformer, utilizing a masked self-attention mechanism. The initial series of GPT models [17, 18] are representative, predicting the next word based solely on preceding words, thus facilitating unidirectional modeling. Seq2seq models such as BART [19] have been developed, leveraging the seq2seq architecture while being specifically tailored for designated tasks.

## 2.2. Local outlier factor

LOF is a density-based outlier detection method [20] and is employed in this study to quantify the novelty of research proposals. This technique identifies local outliers by assessing the degree of isolation of an object relative to its neighboring data points. By focusing on local information, which is often overlooked in traditional methods, LOF effectively captures nuances, making it applicable to various contexts [21-23]. As illustrated in Figure 2, the LOF methodology comprises four essential steps: (1) determining the k-value to initiate the cluster size, (2) calculating the reachability distance for each data point, (3) calculating the local reachability distance value, and (4) generating the LOF score for each data point.

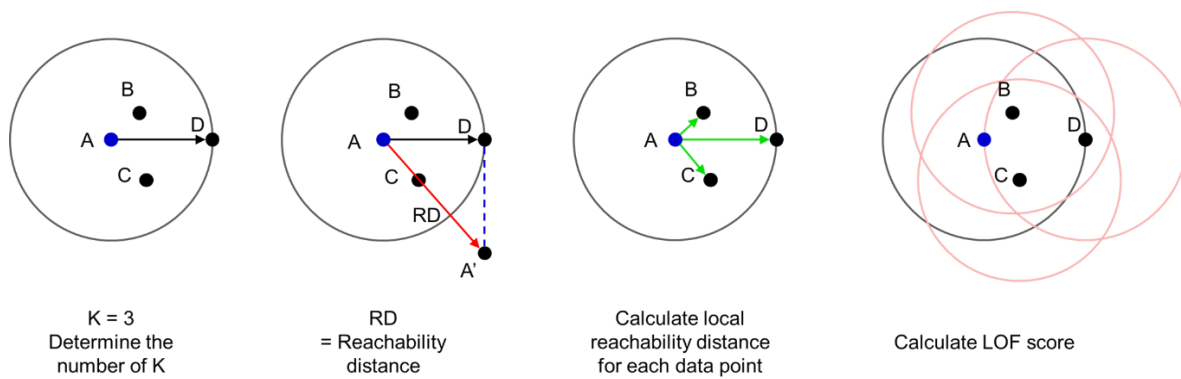

**Figure 2.** Graphical illustration of LOF process

First, the Euclidean distance to the k-th nearest neighbor of an object p is calculated, referred to as the $k - \text{distance}(p)$. Parameter k defines the number of nearest neighbors and can be adjusted based on the analysis requirements. Using this distance, the set of k-nearest neighbors, denoted as kNN(p), consists of all objects within the k-distance from p. Subsequently, the reachability distance from p to object o within kNN(p) is computed as follows:

$$\text{reachDist}_k(p, o) = \max(k - \text{distance}(o, d(p, o)) \qquad \text{Equation (1)}$$

where d(p, o) is the Euclidean distance between p and o. The local reachability density ($lrd_k(p)$) of p is calculated as follows:

$$lrd_k(p) = \frac{k}{\sum_{o \in kNN(p)} reachDist_k(p, o)} \qquad \text{Equation (2)}$$

Finally, the LOF of p for k surrounding neighbors is calculated as follows:

$$\text{LOF(p)} = \frac{1/k \sum_{o \in kNN(p)} lrd_k(o)}{lrd_k(p)} \qquad \text{Equation (3)}$$

In Equation (3), the LOF of p, represented as LOF(p), is defined as the ratio of the average density of p's k-nearest neighbors (kNN(p)) to the density of p. If p is an inlier, its LOF value is approximately 1 because the densities in both the numerator and denominator are comparable. Conversely, if p is an outlier, its LOF value will exceed 1 because of its much lower relative density compared with that of its neighbors. Consequently, an object p that is distant from other objects has a high LOF value, marking it as a potential outlier. The three cases illustrate the LOF computation. First, when p is situated in a dense region, both the local reachability density ($lrd_k(p)$) and that of its neighbors ($lrd_k(o)$) are high, resulting in a low LOF value. Second, if p lies in a uniformly sparse area, both $lrd_k(p)$ and $lrd_k(o)$ are low, thus producing a low LOF value. Finally, when p is in a sparse region surrounded by dense clusters, $\text{lrd}_k(o)$ is high, whereas $(\text{lrd}_k(p))$ is relatively low, leading to a high LOF value that identifies p as an outlier.

## 3. Methodology

The overall process of the proposed approach is described, and a brief description of each step is provided (Figure 3). The proposed approach consists of four steps: (1) data collection and preprocessing, (2) constructing an R&D landscape, (3) measuring the novelty of R&D proposals, and (4) validation.

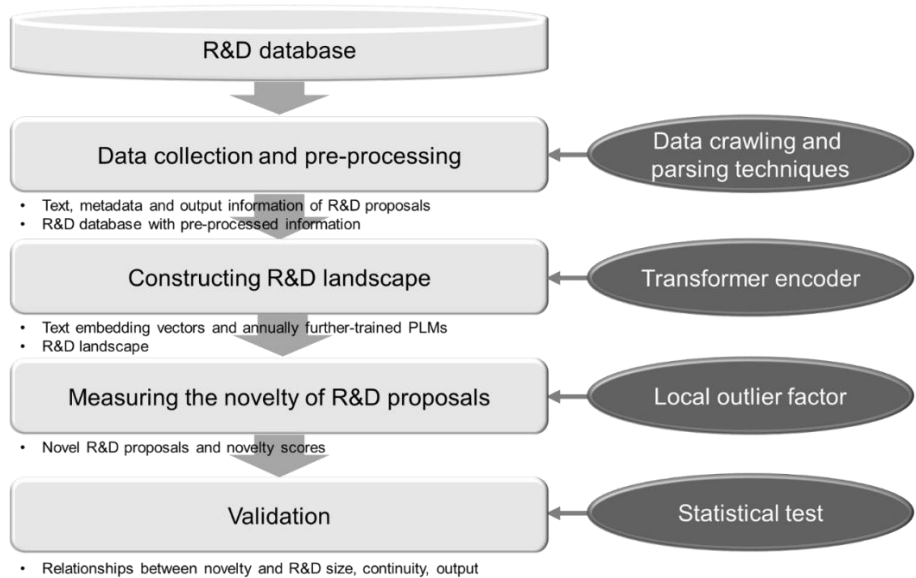

**Figure 1.** Overall process of the proposed approach

### 3.1. Data collection and pre-processing
This step involves collecting and preprocessing research proposals for novelty assessment.

The focus is on proposals selected for funding, as these represent R&D content that has undergone preliminary verification during the review process. In South Korea, such R&D data are managed systematically by the National Science & Technology Information Service (NTIS), which provides comprehensive information on R&D projects and their outcomes, including patents, publications, and technology transfers.

R&D data are generally divided into two main categories: project and performance information. Project information encompasses details provided at the proposal stage, such as project title, research objectives, research contents, expected outcomes, project scale, and institutional affiliations. These textual data often require preprocessing, including the removal of stop words and special characters. Project metadata, including classification codes, is also commonly included; if absent, natural language processing techniques such as topic modeling can be an alternative. In South Korea, diverse classification systems, including the Standard Classification of Science and Technology, have been maintained by the Korea Institute of Science and Technology Evaluation and Planning. An essential attribute in project information is 'continuation status,' which enables the identification of newly selected projects for novelty assessment each year. For projects marked as continuations in year $t$, only the original information submitted as new proposals is used for novelty evaluation. Finally, R&D performance information includes the outcomes associated with each project, recording key achievements such as patents, publications, and technology transfers, which are essential for tracking and assessing R&D impact and contribution.

### 3.2. Constructing R&D landscape

This stage involves constructing an R&D landscape by mapping research proposals into a vector space using a transformer encoder. The annual landscape provides a basis for tracking and analyzing shifts in R&D focus and trends over time. Transformer encoders in the form of PLMs are used to transform textual data into fixed-dimensional embedding vectors, which are critical for structuring and comparing these proposals. Based on language requirements or domain specificity, an appropriate PLM is selected" language-adapted (Multilingual-BERT [24]) and domain-specific (e.g., SciBERT [25], BioBERT [26]).

Although PLMs can be applied directly to downstream tasks, further training, referred to as domain adaptation, may be necessary to capture the specific terminology and contextual nuances within the dataset. This additional pre-training utilizes BERT's sub-word tokenization, which allows the model to learn complex terms by breaking them into sub-words. This approach effectively addresses out-of-vocabulary issues because new terms can be constructed from subword combinations. In summary, domain adaptation can be achieved through further training of a pretrained model on domain-specific data, enabling it to perform downstream tasks such as embedding generation, classification, and prediction with enhanced relevance [27]. In this approach, research proposals are accumulated annually to iteratively refine the PLM, increasing the comprehension of R&D data while incorporating year-over-year changes. This cumulative adaptation allows the model to better represent the evolving landscape of R&D.

### 3.3. Measuring novelty of R&D proposals
In this step, the primary unit of analysis is the research proposal, which can be segmented into multiple components as previously noted. Novelty assessment can be performed on individual components or across an entire proposal. When calculated by component, an LOF score is generated for each, normalized, and averaged to produce an overall novelty measure. LOF input comprises embedding vectors derived from the last layer of PLMs, with dimensionality reduction methods, such as principal component analysis, applied as needed to optimize computational efficiency. A critical parameter in LOF calculation is the number of nearest neighbors $k$, a value generally selected by domain experts. Here, $k$ represents the count of relatively similar proposals within the embedding vector space, determined by either statistically driven or qualitative approaches. Considering that this parameter may be influenced by the desired scope and granularity of the R&D landscape, the selection of $k$ can be flexible. For instance, a higher $k$ value provides a broader neighborhood, suitable for exploring novelty from a broad R&D landscape, facilitating practitioners' macroscopic monitoring. In contrast, a smaller $k$ value increases sensitivity to local density variations, enabling a focused, micro-level view that is valuable for detailed monitoring of novel research proposals within relatively small R&D landscape.

### 3.4 Validation

When introducing a new approach, verifying the validity and quality of its outcomes is essential before its implementation in practice. This study also requires confirmation that the novelty score assigned to research proposals can effectively identify genuinely novel R&D documents. However, given the relative and subjective nature of novelty, absolute verification poses several challenges. To address this, we analyze the relationship between novelty and lagging indicators of R&D performance based on the assumption that novelty is a defining feature of innovation [14], and that innovative R&D often correlates with successful outcomes [28]. The R&D database provides valuable project information, including total research funding allocated to newly selected projects, as well as data on project duration and continuity. Research proposals with high novelty are more likely to receive ongoing R&D support, which reflects their perceived value. In addition, the R&D database records outcomes such as patents, scientific publications, and technology transfers, allowing us to determine whether highly novel research proposals achieve more impactful results than less novel proposals.

## 4. Empirical analysis and results

We conducted a case study of the energy/resource sector R&D in South Korea for the following reasons. First, South Korea's heavy reliance on imported energy resources makes energy security and supply stability vital national priorities [29, 30]. In this context, assessing the originality of research proposals is essential for identifying and funding unique R&D projects, effectively avoiding redundancy and fostering innovation. Furthermore, as global efforts toward energy transition accelerate to mitigate climate change and reduce carbon emissions, South Korea has committed to a 2050 carbon neutrality target, demanding urgent policy reforms and technological advancements [31]. Achieving these goals requires innovative R&D solutions tailored to support sustainable transformation. Finally, the domestic energy policy landscape is evolving rapidly with active discussions on renewable energy expansion, energy efficiency improvements, and nuclear energy policies [32]. In light of these dynamic changes, ensuring the originality of research in the energy and resources sector is critical for enhancing policy alignment and practical viability, reinforcing the ability to navigate and lead in a shifting global energy context.

### 4.1. R&D dataset

We employed NTIS to collect information on R&D project proposals selected between 2010 and 2022. Next, we filtered for proposals ($N = 12,243$) in the Energy/Resources field using the science and technology standard classification designated by KISTEP. The statistics of the collected data are listed in Table 1. The number of new projects has increased since the year 2018. Over the past five years, more than 2,500 R&D projects have been conducted annually in this field, indicating a high level of national interest.

**Table 1.** Summary of collected R&D proposals

| Year | Number of new proposals | Average funding of new proposals (unit: KRW) | Number of total proposals (including ongoing projects) | Average funding of total proposals (unit: KRW) |
|---|---|---|---|---|
| 2010 | 1,052 | 658,503,054 | 1,815 | 740,824,186 |
| 2011 | 940 | 702,677,907 | 1,978 | 693,670,593 |
| 2012 | 990 | 550,630,977 | 2,074 | 758,356,143 |
| 2013 | 894 | 613,881,439 | 2,045 | 933,217,223 |
| 2014 | 863 | 528,182,134 | 2,074 | 932,569,934 |
| 2015 | 898 | 529,530,499 | 2,012 | 788,517,155 |
| 2016 | 728 | 554,132,675 | 1,908 | 681,957,069 |
| 2017 | 1008 | 451,869,410 | 2,144 | 578,247,657 |
| 2018 | 830 | 360,219,932 | 2,510 | 579,133,334 |
| 2019 | 844 | 285,500,815 | 2,741 | 492,702,664 |
| 2020 | 945 | 480,941,773 | 2,613 | 595,079,730 |
| 2021 | 1,127 | 520,049,281 | 2,759 | 683,138,383 |
| 2022 | 1,137 | 488,255,303 | 2,942 | 674,776,867 |

**4.2. R&D landscaping**

In this step, we utilized Korean PLM such as KoBERT ('skt/kobert-base-v1'), which were trained with five million sentences from Korean Wikipedia and 20 million sentences from Korean news. Proposal data were systematically accumulated annually to facilitate further training of the corresponding PLM. The text of proposal title, research objectives, research content, and expected outcomes were concatenated into a single text. As a result, 13 models were developed, corresponding to each year from 2010 to 2022. Table 2 presents partial results of the R&D landscaping analysis derived from the application of the further-trained model for the year 2022 to the research content sections of each proposal. The document was represented as 756-dimensional vectors, and these vectors were used to measure the annual novelty.

**Table 2.** Partial results of R&D landscaping

| Document number | Vector | | | | | | |
| --- | --- | --- | --- | --- | --- | --- | --- |
| | $v_1$ | $v_2$ | $v_3$ | … | $v_{754}$ | $v_{755}$ | $v_{756}$ |
| 1415111320 | -0.1261 | -0.3548 | 0.2060 | … | 0.3822 | -0.4238 | -0.0295 |
| 1425061351 | -0.3393 | -0.2604 | 0.1780 | … | 0.4617 | -0.3012 | -0.1653 |
| 1345135833 | -0.3597 | -0.2294 | 0.1890 | … | 0.4028 | -0.2576 | -0.1309 |
| 1345135814 | -0.3821 | -0.3011 | 0.0881 | … | 0.4680 | -0.2896 | -0.0870 |
| 1425065831 | -0.3468 | -0.2721 | 0.2130 | … | 0.4713 | -0.2847 | -0.1926 |
| … | … | … | … | … | … | … | … |
| 1425166910 | -0.3456 | -0.2188 | 0.1674 | … | 0.4527 | -0.2633 | -0.1878 |
| 1711158208 | -0.3364 | -0.2586 | 0.2262 | … | 0.4407 | -0.2981 | -0.1595 |
| 1345353596 | -.3576 | -0.2481 | 0.2000 | … | -0.4138 | -0.2731 | -0.1948 |
| 1345354020 | -0.2904 | -0.1905 | 0.1681 | … | 0.4146 | -0.2824 | -0.1225 |
| 1345354195 | -0.3059 | -0.1764 | 0.1488 | … | 0.4282 | -0.2673 | -0.0732 |

**Note:** The text used is part of the 'research contents' section in a research proposal, and the embedding vectors were generated using a KoBERT-based transformer model, further trained on research proposals up to 2022.

### 4.3. Novelty measurement

The Python library scikit-learn [33] was used to implement the LOF algorithm to assess the novelty of each research proposal document. The number of nearest neighbors k was defined as 1% of the total number of new proposals accumulated annually to flexibly address the increasing R&D landscape over time. LOF was applied to the embeddings of the four key components within each proposal. By calculating LOF scores, proposals with lower local density values received higher LOF scores, facilitating the identification of potentially novel concepts within the proposal set. Table 3 presents a subset of novelty assessment results, showing LOF scores that were calculated and normalized for each year.

**Table 3.** Partial results of novelty measurement

| Document number | Year | Novelty of 'proposal title' | Novelty of 'research objectives' | Novelty of 'research contents' | Novelty of 'expected outcomes' | Total novelty |
|---|---|---|---|---|---|---|
| 1415111320 | 2010 | 0.1960 | 0.3494 | 0.3700 | 0.8704 | 0.4464 |
| 1425061351 | 2010 | 0.4174 | 0.1570 | 0.3993 | 0.7658 | 0.4349 |
| 1345135833 | 2010 | 0.1007 | 0.1921 | 0.3835 | 1.0000 | 0.4191 |
| 1345135814 | 2010 | 0.9073 | 0.4894 | 0.0427 | 0.0762 | 0.3789 |

| 1425065831 | 2010 | 0.2246 | 1.0000 | 0.0847 | 0.0399 | 0.3373 |
| --- | --- | --- | --- | --- | --- | --- |
| … | … | … | … | … | … | … |
| 1425166910 | 2022 | 0.3834 | 0.5246 | 0.0179 | 0.0302 | 0.2390 |
| 1711158208 | 2022 | 0.5331 | 0.0251 | 0.0620 | 0.2830 | 0.2258 |
| 1345353596 | 2022 | 0.0878 | 0.2642 | 0.1924 | 0.3321 | 0.2191 |
| 1345354020 | 2022 | 0.0687 | 0.2396 | 0.1928 | 0.3573 | 0.2146 |
| 1345354195 | 2022 | 0.1147 | 0.2436 | 0.1967 | 0.3006 | 0.2139 |

**Note:** The number of nearest neighbours k for each year is determined as follows; 2010: 10, 2011: 20, 2012: 30, 2013: 39, 2014: 47, 2015: 56, 2016: 64, 2017: 74, 2018: 82, 2019: 90, 2020: 100, 2021: 111, 2022: 122

In the 2010 cohort of newly selected proposals, those with the highest novelty scores included proposal number 1415111320, "Development of high-performance thermoelectric composites and sputtering targets by spark plasma sintering," and proposal number 1425061351, "Development of solar cell wafer etching equipment," achieving novelty scores of 0.4464 and 0.4349, respectively. Proposal 1345135833, "Development of chameleon windows with energy storage capabilities and application to sustainable building structures," exhibited particularly high novelty in the expected outcomes category. Conversely, proposal 1345135814, "Development of high-efficiency biofuels and low-temperature μ-SOFC using 3D nanostructure networks," displayed high novelty in its title, underscoring the innovative framework introduced. In 2022, notable novelty scores were recorded for proposal number 1425166910, "1 kW-class portable hydrogen fuel cell generator using low-pressure hydrogen storage," and proposal number 1711158208, "Atomic-level surface control technology for electrochemical complex oxide materials for energy conversion," scoring 0.2390 and 0.2258, respectively. Interestingly, proposal 1345353596, "Development of high-entropy multi-ion metal catalysts for hydrogen generation," showed exceptionally high novelty within the expected outcome category, indicating its high industrial implications. By calculating relative novelty on an annual basis, we can track how each proposal's novelty is assessed over time (Table 4). Proposals with high novelty scores in 2010, for example, tended to exhibit decreased relative novelty in subsequent years. This decline highlights an evolving research landscape in which the perceived novelty of certain topics shifts as new advancements and research trends emerge. These variations in novelty scores underscore the dynamic nature of R&D, suggesting that the context and criteria for evaluating novelty adapt in response to ongoing developments in the field.

**Table 4.** Partial result of annual measurement of total novelty scores

| Document number | Year | | | | | | |
|---|---|---|---|---|---|---|---|
| | 2010 | 2011 | … | 2016 | 2017 | … | 2022 |
| 1415111320 | 0.4464 | 0.4804 | … | 0.2228 | 0.2392 | … | 0.3355 |
| 1425061351 | 0.4349 | 0.2582 | … | 0.2791 | 0.1830 | … | 0.2855 |
| 1345135833 | 0.4191 | 0.3363 | … | 0.2791 | 0.0756 | … | 0.1975 |
| 1345135814 | 0.3789 | 0.0663 | … | 0.096 | 0.0405 | … | 0.0905 |
| 1425065831 | 0.3373 | 0.0928 | … | 0.0989 | 0.0709 | … | 0.1200 |
| … | … | … | … | … | … | … | … |
| 1425166910 | − | − | − | − | − | − | 0.2390 |
| 1711158208 | − | − | − | − | − | − | 0.2258 |
| 1345353596 | − | − | − | − | − | − | 0.2191 |
| 1345354020 | − | − | − | − | − | − | 0.2146 |
| 1345354195 | − | − | − | − | − | − | 0.2139 |

**Note**: The novelty scores for newly selected proposals in 2022 are evaluated solely within that year, meaning no novelty scores exist for prior or subsequent years. By contrast, proposals selected in 2010 continue to receive updated novelty scores each year as new documents enter the landscape and undergo novelty evaluation.

### 4.3. Validation

After calculating the annual novelty scores of newly submitted research proposals, we designated the top 10% with the highest scores as novel proposals, with the remaining categorized as non-novel. We then conducted a statistical analysis to determine whether these novel proposals demonstrated superior performance in terms of R&D continuity, size, and outputs, such as patents, publications, and technology transfers. Specifically, we assessed project duration; initial total funding; and the number of domestic/foreign patents, publications, and technology transfers produced. The Mann-Whitney U test [34] was applied to these two groups, providing a non-parametric method to evaluate differences in medians given that the data did not meet the normality assumption required for parametric tests. Here, patents refer to granted patents, and publications are limited to those that appear in SCI-indexed journals. We selectively employed proposals from 2010 to 2020 (7,360 proposals), as more recent proposals lack sufficient data to allow for robust comparison of these lagging indicators across groups.

As summarized in Table 5, novel research proposals typically have shorter project durations, a higher probability of technology transfer, and lower counts of SCI-indexed publications than non-novel proposals. The p-values for the indicators of R&D continuity and R&D output (publications and technology transfer) were 0.0046, 0.0001, and 0.0066, respectively, below the significance threshold of 0.05, indicating statistically significant differences between novel and non-novel proposals for these metrics. These findings suggest that while novel projects often involve high levels of creativity, they may also face greater uncertainty, potentially resulting in shorter project lifespans. In addition, novel projects may encounter challenges in the journal review process because of their limited alignment with existing research, which may affect their publication counts. Interestingly, novel proposals are significantly more likely to result in technology transfer than non-novel ones, suggesting that successful novel research can have substantial impacts, particularly given the relative rarity of technology transfer as an outcome. By contrast, the analysis of R&D project size (total funding at project initiation) yielded a p-value of 0.4908, which is considerably higher than the significance level of 0.05, indicating no statistically significant difference in initial funding between the two groups.

**Table 5.** Results of validating the proposed approach for dataset before 2021

| Category | | Novel proposals | Non-novel proposals |
|---|---|---|---|
| Number of observations | | 912 | 9067 |
| R&D continuity | Mean | 2.2138 | 2.3597 |
| | S.D. | 1.2331 | 1.3426 |
| R&D output (papers) | Mean | 3.0154 | 4.3629 |
| | S.D. | 9.5623 | 15.9849 |
| R&D output (technology transfers) | Mean | 0.8542 | 0.7724 |
| | S.D. | 4.1876 | 7.6599 |

| R&D output* (domestic patents) | Mean | 1.7478 | 1.9333 |
| --- | --- | --- | --- |
| | S.D. | 4.7845 | 5.1108 |
| R&D output* (foreign patents) | Mean | 0.1447 | 0.1625 |
| | S.D. | 1.0471 | 2.3919 |

**Note:** The p-values for domestic and foreign patent counts were slightly higher at 0.0707 and 0.1383, suggesting that the difference in patent counts between the two groups may not be statistically significant.

## 5. Discussion
### 5.1. Implications for theory and practice

The proposed approach presents a novel framework for constructing and navigating complex and rapidly evolving R&D landscapes, offering substantial academic and practical implications. First, this methodology is designed as a highly replicable and adaptable structure that enables experts to construct, explore, and interpret R&D landscapes using data from proposals that contain near-future R&D plans. This provides new theoretical insights into the literature on R&D planning and monitoring. In contrast to prior studies that largely relied on patents or publications as R&D outputs to map landscapes, our approach leveraged forward-looking proposal data to build and explore future-oriented R&D landscapes. Given the complex and specialized content of research proposals, we employed a transformer-based language model to ensure a precise and comprehensive understanding of R&D documentation. To our knowledge, this study is the first attempt to develop a language model specific to R&D proposals to construct a detailed R&D landscape. Our language model-based analysis of R&D proposal texts advances beyond keyword-level monitoring to facilitate extensive comprehension of scientific and technological insights. Therefore, this proposed approach offers policymakers and R&D managers a strategic tool for forward-thinking R&D planning and decision-making, presenting domain-specific and quantitative novelty indicators within the R&D landscape. Moreover, the systematic procedures and rigorous methodologies established in this study enable the development of quantitative metrics to assess novelty in R&D proposals. Although our primary research focus was on identifying novel R&D proposals, our approach and findings have broader applications, including forecasting future R&D trends, detecting emerging weak signals, and analyzing open innovation ecosystems.

Second, the proposed approach offers substantial practical implications by enabling the development of automated software systems that enhance big data-driven R&D landscaping for domain experts, even those without specific expertise in AI or natural language processing. The systematic framework is well-suited to software implementation, allowing users to interactively execute individual steps, such as dataset curation, R&D landscape construction, and novelty assessment, receiving real-time feedback and intermediate results tailored to expert input. Once established, the software provides users with end-to-end R&D landscaping and novelty assessments, streamlining the entire process. For users aiming to tailor outcomes to specific research objectives, the system allows for adjustments such as careful data selection and parameter tuning to achieve targeted analysis. In practice, institutions and organizations may also benefit from periodic updates to the R&D landscape and language models, reflecting the latest developments in scientific and technological knowledge and maintaining the relevance of ongoing R&D monitoring. In addition, this approach enables users to flexibly adjust dataset parameters and incorporate new data with ease, facilitating customized and dynamic updates to R&D landscapes aligned with specific research domains or strategic goals. Once a research field of interest is identified, the system can autonomously support further steps, such as filtering proposal data by classification codes, retraining the language model, and reconstructing the R&D landscape as needed to reflect evolving research priorities.

## 5.2. Implementation of the proposed approach

This study suggests a systematic approach for constructing and exploring R&D landscapes with quantitative outcomes and scientific methods. The proposed approach offers several distinctive advantages over traditional methods. First, it allows experts to efficiently identify novel R&D proposals across an expansive R&D landscape, thereby significantly reducing the time and effort required for comprehensive R&D monitoring. Second, the approach serves as a useful decision-support tool for R&D planning, as it highlights proposals with high novelty compared to similar or previous ones, providing experts with critical insights into emerging trends. Through this interactive process, experts can define the scope of the R&D landscape and pinpoint novel proposals based on their specialized knowledge and insights. Such flexibility has substantial implications in real-world applications where varying industry contexts and dynamic conditions necessitate adaptable tools. Third, our approach offers practical advantages in handling large volumes of R&D documentation, which is a common challenge in the decision-making process. By developing a transformer-based model specifically tuned to R&D data, this approach not only improves domain knowledge comprehension but also enhances document processing efficiency. This specialized model allows for more effective management and exploration of R&D landscapes, facilitating precise and insightful analysis of complex R&D data, an essential capability in environments requiring timely and informed decision-making.

Despite these advantages, implementing these newly developed methods in a practical setting requires careful consideration. Potential users should evaluate several key factors when adopting the proposed approach, particularly the following considerations that hold across domains. First, broad exploration of novel R&D proposals is essential, as this study's approach broadens expert perspectives on R&D novelty and provides insights distinct from prior research. In addition, organizations should customize language models based on dataset characteristics (e.g., language, domain, and document type) to enhance model accuracy, with further training recommended for specialized R&D vocabularies. Third, flexibility in defining datasets and novelty criteria is crucial; users can adjust these parameters to meet specific objectives, leveraging techniques such as topic modeling for targeted monitoring [35]. Expert validation and qualitative reviews remain essential, particularly for interpreting results and ensuring reassessment of R&D novelty. Finally, adjustable parameters in the approach, such as embedding dimensions and number of nearest neighbors, allow for fine-tuning that aligns with specific analysis requirements. With these considerations, organizations can effectively tailor their proposed approach to enhance R&D monitoring and novelty detection.

## 6. Conclusion

The strategic role of R&D landscaping is becoming increasingly essential for obtaining precise and actionable insights into today's complex and dynamic R&D ecosystem. This study presents a forward-looking approach to R&D landscaping by systematically identifying novel R&D proposals through rigorous quantitative analysis. This study posits that R&D project proposals provide critical insights into both the historical and prospective dimensions of the R&D landscape. Focusing on novelty as a primary driver of innovation, this study explored new perspectives within the landscape. To this end, a comprehensive R&D landscape is constructed using transformer-based language models that capture new insights from the proposal data. The annual identification of relatively novel proposals was achieved through LOF analysis. The subsequent evaluation of these proposals for R&D continuity and output revealed a strong correlation between novelty and successful technology transfer. A case study of South Korea's energy/resource sector highlights the effectiveness of the proposed approach in enabling a deep, comprehensive exploration of complex R&D landscapes.

This study has several limitations. First, regarding the assessment of novelty in research proposals, our approach primarily measured the semantic dissimilarity of proposals relative to other documents. However, this method cannot definitively confirm the originality of research proposals. The proposed novelty indicator would gain further validity if combined with expert evaluations of project novelty. Second, concerning the scope of novelty analysis, this study treats all sections within the proposals as units for novelty evaluation. However, novelty can manifest at more granular levels, such as sentences, words, or even specific knowledge entities. For example, in the energy/resources sector, novelty may arise from the introduction of new materials to existing challenges or the novel application of established materials. Third, in terms of language modeling, although our approach utilizes transformer-based language models tailored to R&D proposals, opportunities for improvement remain, particularly by integrating state-of-the-art advancements. Although the combined use of LOF analysis and transformers enhances the model's effectiveness, computational complexity may increase with the expansion of R&D data or model scale, which requires considerable computational resources. Fourth, while most stages of our method are automated, certain steps, such as interpreting novel proposals and analyzing causative factors, still depend on expert judgment. Here, attention-based text classification models could assist in identifying novel documents by highlighting specific terms or sentences that contribute most to perceived novelty [36], contingent upon the full validation of novelty. Lastly, this study's reliance on a single case study limits the comprehensive validation of the proposed approach's utility and effectiveness, as the identified novelty and R&D landscape lack direct outcome-based verification and rely instead on indirect evidence.